# VideoMCC: a New Benchmark for Video Comprehension

Du Tran[1,2], Maksim Bolonkin[1], Manohar Paluri[2], Lorenzo Torresani[1]
[1]Dartmouth College, [2]Facebook AI Research
{trandu, mano}@fb.com, {maksim.bolonkin.gr, lt}@dartmouh.edu

## Abstract

*While there is overall agreement that future technology for organizing, browsing and searching videos hinges on the development of methods for high-level semantic understanding of video, so far no consensus has been reached on the best way to train and assess models for this task. Casting video understanding as a form of action or event categorization is difficult as it is not fully clear what the semantic classes or abstractions in this domain should be. Language allows to sidestep the problem of defining video categories, by formulating video understanding as the task of captioning or description. However, language is redundant and sometimes ambiguous. Many different captions may express the same semantic concept. To account for this ambiguity, quantitative evaluation of video description requires sophisticated metrics, whose performance scores are hard to interpret by humans.*

*This paper provides four contributions to this problem. First, we formulate* Video Multiple Choice Caption (VideoMCC) *as a way of assessing video comprehension through an easy-to-interpret performance measure. Second, we describe a general semi-automatic procedure to create benchmarks for this task. Third, we publicly release a large-scale video benchmark created with an implementation of this procedure and we include a human study that assesses human performance on our dataset. Finally, we propose and test a varied collection of approaches on this benchmark for the purpose of gaining a better understanding of the new challenges posed by video comprehension.*

## 1. Introduction

Over the last few years deep learning has revolutionized the field of still-image analysis by delivering breakthrough results on object categorization [34, 23], detection [16], scene classification [46], and semantic segmentation [26]. These successes were driven in large part by the introduction of large scale datasets [13, 18, 47] that made it possible to train effectively deep image models with large learning capacity.

Unfortunately in the video domain many analysis benchmarks (e.g., HMDB51 [24], UCF101 [38]) are still too small in size to enable effective learning of deep models. Furthermore, the labels manually collected on these datasets merely specify the class of the action in each video (e.g., *walking* or *sitting*) but do not indicate where the action is performed. Thus, the learning algorithm is left with the burden of discovering on its own the portion of the video that is truly representative of the action. Some of these datasets have limited semantic scope, as they either include a small number of action categories (e.g., 101 classes for UCF101) or are focused on specific domains (e.g., just sport activities for the case of Sports-1M [19]). Therefore, features learned from such datasets are unlikely to perform well on videos containing more general, everyday actions.

For these reasons, several authors [44, 41, 33] have proposed to reformulate video understanding as a description or captioning task, where the goal is to describe the input video in text form. A benefit of this is that such output is directly readable by humans. The downside however is that these outputs are hard to evaluate quantitatively. Since many different sentences can reasonably describe a given video, for each input there are many correct outputs. To address this ambiguity, one can resort to comparative evaluation by human judges [11]. But this would require a huge crowd-sourcing effort for every new algorithm to assess. Another approach is to design sophisticated metrics (e.g., METEOR or BLUE) that can capture the similarities of captions expressing the same semantic concept. However, it is hard for humans to interpret the meaning of these scores. E.g., is a METEOR score of 28% representing an acceptable captioning performance, or how much difference in these scores would lead to a noticeable difference in predictions?

In this paper we propose to cast video understanding in the form of multiple choice tests that assess the ability of the algorithm to comprehend the semantics of the video. Figure 1 illustrates an example of such test. The algorithm is presented with an input video and $k$ possible descriptions, where $k-1$ of them are distractors. The method must choose the description that best matches the video. The task is well-posed as a traditional classification problem, with performance numbers easy to interpret (e.g., random chance

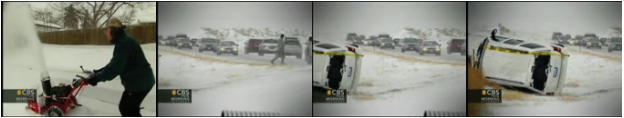

A. It was not after the second World War.
B. We really want to step up to the plate and be a part of the problem solving situation.
C. That only happens because there's demand for our product.
D. 600 flights canceled, as much as three feet of snow with five-foot drifts.
E. Thank you for joining us.

Figure 1: **VideoMCC example**. Video Multiple Choice Caption requires choosing one of $k$ possible sentences as the description for an input video clip. In this example the correct answer is (D).

produces an accuracy of $1/k$). Yet, this classification task does not require the definition of video classes or action categories. Furthermore, we describe a procedure to construct multiple-choice tests with very little human intervention. This makes it possible to generate large-scale benchmarks for training and testing deep models on this task. Using this procedure we built a dataset that we will release to the research community. Although this is only the first version of our benchmark, it has already size comparable to the largest existing datasets for video analysis. Finally, in this paper we also present preliminary results achieved with several approaches to video comprehension, including regression and metric learning. In summary, the contributions of our work are four-fold:

1. We propose to cast high-level video understanding in the form of multiple choice tests. This task is well-posed and easy to evaluate.
2. We describe a general semi-automatic procedure to construct benchmarks for this task (section 3.2). The procedure requires a small set of manual annotations, independent of the size of the dataset. This renders our approach suitable to build benchmarks of unprecedented scale.
3. We present an implementation of this procedure, which we used to create a video benchmark of size comparable to the biggest existing datasets (section 3.3).
4. We introduce and assess a varied set of simple methods to tackle the problem of VideoMCC on our benchmark (section 4).

## 2. Related Work

Video understanding has been studied for many years. Early approaches focused on action recognition [14, 6, 38], event detection [20], irregularity detection [8], action similarity labeling [22]. Most of these methods rely on handcrafted features [25, 45] and train machine leaning models on top of these representations. Recent advances in deep learning have opened up the possibility of learning models from raw videos. Simonyan and Zisserman introduced a two-stream network that achieved strong results on action recognition [36]. Tran *et. al.* proposed to use 3D ConvNets to learn spatiotemporal features from a large-scale dataset [43]. Despite their good performance on action categorization, these approaches are by design limited to predict a single label per video and thus provide a very coarse characterization of the semantics in the video.

Inspired by recent promising results on image captioning [11], different approaches have been proposed for video description [44, 41]. Several of these methods are based on recurrent neural networks (e.g., LSTM) and are trained to predict a sentence describing the input video. This area shows good promise for developing algorithms that can understand and describe videos in a human-readable language. However, captioning is very hard to assess. This limitation makes it hard to compare competing algorithms, even when resorting to human judges. Visual question and answer (QA) was also recently introduced for both images [5] and videos [39]. Compared to captioning, Visual QA is better-posed, as the problem is conditioned on the question being asked. However, in the free-form QA setting, there are still multiple correct answers that can be correctly applied to a single question. Moreover, collecting ground truth annotations for QA is very expensive. This makes it hard to build large-scale datasets on this task.

The variation of video comprehension task that we propose shares similarities with video captioning and QA, as it also assesses algorithms on their ability to understand the semantics of the video. However, our task is different in its formulation: it entails selecting one of the $k$ possible Closed Captions (CC) from a multiple choice test, rather than asking to describe or to answer a question. This renders the quantitative evaluation easy to carry out and makes performance scores very intuitive. While one may argue that CCs do not always provide a *complete* description of the video, we point out that the same can be said about video description datasets, since their annotations typically focus on the most salient visual aspects, as determined by the human annotator. Our task merely requires finding the CC that best agrees with the video. We contend that this goal is well defined even when the true CC does not provide a complete description of the video.

Finally, we note that while any video description dataset can in principle be turned into a multiple-choice dataset similar to ours, all existing video description benchmarks are based on manual annotations and, as a result of this large human cost, they are either small (see Table 1) or prohibitive to scale to bigger sizes. For example, in unpublished work concurrent to ours, Atousa et al. [42] propose a multiple-choice video benchmark built on the existing LSDMC16 dataset [33]. While this dataset has fairly large size (118K clips, 158 hours), it is difficult to scale further, as it relies on movie audio descriptions, which are prepared by trained professional describers, require up to 60 person-hours for a 2-hour movie, and are currently available only for a small subset of movies. Similarly, in another unpublished work, Zhu et al. [48] present a multiple-choice benchmark involving about 110K video clips obtained from an aggregation of three existing manually-annotated datasets [30, 31, 3]. Unlike all these prior efforts, our semi-automatic procedure does not require any human intervention in the annotation of the videos, except for a small training set needed for learning the relevant-clip detector. But the size of this training set is uncorrelated to the size of the video dataset, which can be grown arbitrarily large without any further human labeling.

We stress that we do not claim that VideoMCC is "better" than captioning or action recognition but rather that it is *another* valuable proxy for video understanding and that it offers the added benefits of ease of scalability, content diversity, and a performance metric that is easy to interpret and to evaluate.

## 3. Video Multiple Choice Caption

### 3.1. Problem statement

Given an input video clip $\mathcal{V}$ and a set of $k$ text sentences $s_1, s_2, \ldots, s_k$, the problem of video comprehension is to predict which of these $k$ choices best describes the visual content of the input clip. Note that readable text in the frames is automatically blurred and audio is removed. A concrete example of video comprehension is provided in Figure 1. We argue that for a system to do well on this task it must be able to infer the true semantics of the video, including context, the nature of the interactions among the subjects, and the objects appearing in the scene. Thus, it represents a good assessment of video comprehension by machines.

### 3.2. Constructing a VideoMCC Dataset

In order to assess and compare methods on the task of VideoMCC, a dataset must be constructed to enable the training and testing of models on this problem. Here we review the desiderata that inspired the construction design of our benchmark. Ideally, the dataset must be:

1. **Large-scale**. In the still-image domain we have witnessed a revolution in methodology and dramatic impovements with the introduction of a large-scale dataset [34]. In fact, recent research has shown that the problems of overfitting and difficult optimization with deep models are vastly reduced when training on large datasets. Thus, our desired benchmark should be large enough to enable effective training of these models.
2. **Semi-automatic**. The process of dataset construction must be semi-automatic and must require little human intervention. This is a fundamental requirement in order to build a massive collection of examples. We note that the limited scale of prior datasets in the video domain is a direct consequence of the high human cost and time consumption needed to label video clips.
3. **Semantically diverse**. As our goal is to train models that can comprehend video of arbitrary nature, the dataset must contain a wide representation of subjects, including politics, sports, science, technology, and arts.

To meet these criteria, we propose a procedure that generates semi-automatically VideoMCC tests (as shown in Figure 1) by leveraging an existing repository of TV news programs – the TV News Archive [2]. We note that access to the Archive's Collections is granted at no cost for scholarship and research. Thus, it represents a fitting platform for the construction of video benchmarks. Furthermore, as TV news cover all social, cultural, and natural aspects of modern life, the collection is inherently semantically diverse. Finally, the videos have associated time-synchronized English captions providing a well-aligned textual transcription of the audio (the TV News Archive uses sphinx and phonemes to align the timing of the broadcasted captions with video). We utilize these closed captions (CCs) to automatically generate the textual descriptions corresponding to the ground truth labels of the videos. This source of information allows our procedure to generate a massive collection of comprehension tests with ground truth labels almost fully automatically (as further explained below, a small set of initial human annotations are needed to bootstrap the process).

We now discuss in detail the construction of the dataset. Each video downloaded from the Archive is a complete TV news show from a particular channel broadcasted on a specific day (e.g. *ABC News Good Morning America* on August 27, 2011 from 8am to 9am) with lengths varying from 30 minutes to 2 hours. Our procedure then performs a sequence of steps aimed at generating a set of video comprehension tests from each program. The steps include clip segmentation, clip elimination, and multiple-choice test generation.

**Clip segmentation**. Each TV news video is segmented into short clips, corresponding to individual sentences (terminated by a period) of the CCs. For each sentence, using the time stamps of its start and end, we segment the corresponding clip from the video. This process yields a massive number of clips. In order to build a dataset of clips having fairly homogeneous lengths and to have enough temporal context for each clip, we eliminate clips that are shorter than 2 seconds or longer than 5 seconds. Similarly, we discard clips corresponding to sentences that are too short (fewer than 5 words) or too long (more than 60 words).

**Clip elimination**. This step is carried out to remove clips whose visual content is not informative. Examples include advertisement, static scenes, segments showing anchors speaking, sections inside the news studio, such as the weather forecast portion of the news program. Such clips are not useful for training general computer vision models. In order to make our dataset construction scalable, we develop a detector to automatically discard irrelevant clips. The detector is trained on a small collection of clips manu-ally labeled as either *irrelevant* (e.g. studio, advertisement, weather-forecast clips) or *relevant* (out-of-studio footage, such as dynamic scenes where human subjects or the camera are moving). The detector is trained on the visual component of each clip (thus, without considering its CC). The details of our detector are discussed in section 3.3.

| Dataset | UCF101 [38] | Youtube-8M [4] | Youtube-BB [29] | Sports1M [19] | ActivityNet [15] | MSVD [10] | MPII-MD [32] | M-VAD [41] | MovieQA [39] | VideoMCC (ours) |
|---|---|---|---|---|---|---|---|---|---|---|
| # clips | 13K | 8M | 380K | 1.1M | 28K | 2K | 68K | 49K | 15K | 272K |
| # hours | 24 | 500K | N/A | N/A | 648 | 5 | 74 | 85 | N/A | 628 |
| Task | cls | cls | det | cls | cls/det | des | des | des | qa | videomcc |
| Type | action | activity/objects | objects | sport | action | movie | movie | movie | movie | news |

Table 1: **Dataset comparison**. Comparison of video datasets in term of size and task. Abbreviations for tasks: cls (classification), det (detection), des (description), qa (question and answer), and videomcc (Video Multiple Choice Caption). N/A: information is not available in [29, 19, 39].

**Multiple-choice test generation**. Given a video clip, we form a multiple-choice test of $k$ potential textual descriptions by including $k-1$ distractors and the true associated CC sentence. The distractors can be selected in different ways. The simplest solution is to randomly sample the $k-1$ distractors from the entire set of CC sentences. In order to make the test more challenging, one may want to select distractors that are not too distant from the correct response, according to a semantic metric over text descriptions, such as the distance of word2vec vectors representing sentences [28].

### 3.3. The VideoMCC Dataset

In this section we discuss a specific implementation of the general procedure outlined above. This implementation was used to build a dataset of $272,504$ VideoMCC tests, which we will make publicly available to the research community. The benchmark is split into $191,028$ training examples and $81,476$ test examples. The dataset is constructed from $4,990$ news videos from the TV News Archive. These videos were obtained by uniformly sampling 77 distinct daily TV news shows (BBC World News, MSNBC News Live, PBS News Hour, etc.) in the period from January 1, 2009 to December 31, 2014.

We use a subset of 20 news videos (randomly selected from our original $4,990$ TV videos) as a training set exclusively for the development of our relevant/irrelevant clip detector. Note that we remove these 20 videos from the collection used for dataset construction. We manually labeled all clips segmented from this set as either irrelevant (e.g. studio, advertisement, weather-forecast clips) or relevant. We represent each clip using the C3D spatiotemporal features [43], which are activations of a convolutional neural network (ConvNet) optimized for action classification. We use the activations from layer fc6. We opted for this descriptor as it has been shown by the authors to yield good performance on a variety of tasks involving semantic analysis of video. We train a simple linear SVM on this representation to classify whether a clip is relevant or not. We evaluate this detector on our training set of 20 videos using 20-fold cross validation. The resulting ROC curve is shown in the Appendix B. The detector achieves an area under the curve (AUC) of $0.94$. We use this ROC curve to choose the cutoff threshold to reject irrelevant clips. We chose the threshold corresponding to a false positive rate of $0.1$, which yields a true positive rate of $0.83$. This represents a good trade-off in terms of recall vs error (in other words, to retrieve $83\%$ of the relevant clips we must cope with only $10\%$ of irrelevant clips). Further filtering of these clips could be performed via crowdsourcing at a fairly limited financial cost. However, our experiments suggest that a $10\%$ of irrelevant clips (i.e., clips whose visual content is not strongly correlated with their CCs) in our dataset does not prevent the training of effective models for video comprehension but it offers the big benefit of a semi-automatic solution.

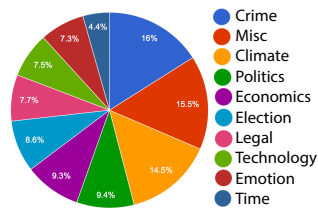

Figure 2: **VideoMCC topic distribution**. The subject distribution was estimated by training an LDA [7] model with 10 topics on closed caption sentences of VideoMCC clips.

Applying this detector to the remaining $4970$ videos yields a total of 272K clips deemed relevant for the purpose of VideoMCC. We partition this dataset into training and testing splits (using a ratio of 7:3), with the additional constraint that all clips from a video are inserted in the same split (either training or testing) to avoid bias.

Table 1 compares VideoMCC to existing video datasets in terms of size, task, and content. VideoMCC is one of the biggest datasets in terms of both number of hours and number of clips. However, while in Sports-1M and Youtube-8M (the largest datasets in this comparison) each clip is labeled with one or more action/entity classes, each clip in VideoMCC is labeled with a textual description, which often provides a semantically richer annotation than a class label.

In order to understand the distribution of subject matters represented in VideoMCC clips, we trained an LDA [7] model on the CC sentences of our entire training set using

10 topics. We visually inspected the most frequent words of each topic in order to manually assign a subject tag to each topic ("politics", "economics", "technology", etc). The most frequent words are listed in a table included in the Appendix A. Figure 2 shows the subject distribution computed on the 272K clips of VideoMCC. It can be seen that the breadth of topics covered in VideoMCC distinguishes our dataset from prior video collections, which are much more narrowly focused in content (e.g., videos depicting only sports or movies). This makes VideoMCC particularly fitting for the training of models for general and comprehensive video understanding.

For each clip we formed a multiple-choice test by randomly selecting 4 distractor sentences from our entire pool of CC sentences, in addition to the correct answer (the true CC). Thus, each clip involves $k = 5$ possible descriptions.

## 4. Approaches to VideoMCC

In this section, we consider two simple approaches for Video Multiple Choice Caption on our dataset. In order to understand well this problem and the dataset, we opt on purpose for a set of simple baselines. While more sophisticated approaches may lead to improved performance, they are beyond the scope of this initial study and are reserved to future work, by us and the rest of the community.

We first introduce our notation. Let us denote the training set with $\{x^i, y_1^i, y_2^i, \ldots, y_k^i, t^i\}_{1..n}$, where $x^i$ is the $i$-th video clip, $y_1^i, y_2^i, \ldots, y_k^i$ are the $k$ sentences defining the multiple choice test, and $t^i \in \{1..k\}$ is the answer key, i.e., the index to the correct answer. Let $\phi_v(x)$ be a visual embedding (i.e., a feature representation computed from pixel values) of video clip $x$ and $\phi_l(y)$ the language embedding of the text sentence $y$. Examples of possible choices for the visual embedding include aggregations of deep image features computed from individual frames of the clip (e.g., average pooling of VGG activations [37]) or deep video clip descriptors (e.g., C3D fc6 activations [43]). The language embedding can be produced, e.g., by averaging the word2vec representation [28] of all words in the sentence.

### 4.1. Regression

A simple strategy to approach VideoMCC is to train a regression model $\mathcal{R}(x; \mathbf{W})$ parameterized by weights $\mathbf{W}$ to map from the the visual embedding to the language embedding, i.e., such that $\mathcal{R}(x^i; \mathbf{W}) \approx \phi_l(y_{t^i}^i)$. A simple instantiation of this method would involve learning a linear transformation of $\phi_v(x)$, i.e., $\mathcal{R}(x^i; \mathbf{W}) = \mathbf{W}\phi_v(x^i)$, where the parameter matrix $\mathbf{W}$ can be estimated via least-squares regression. Predictions can then be made by choosing the sentence whose language descriptor is closest to the transformed visual vector, i.e., $t^* = \operatorname{argmin}_{t \in [1..k]} \|\mathbf{W}\phi_v(x) - \phi_l(y_t)\|_2^2$.

This strategy can be rendered more powerful by replacing the linear regression model with a deep convolutional network $\mathcal{R}(x; \mathbf{W})$ (here $\mathbf{W}$ denotes weigths) that is trained directly on the raw video input $x^i$ to regress the associated CC vector $\phi_l(y_{t^i}^i)$.

### 4.2. Metric learning

It can be argued that the regression strategy outlined above is overly aggressive as it forces the visual vectors to be mapped into their language counterparts. This objective is difficult to realize due to the high-dimensionality of the output space. We can relax this desideratum by merely requiring that the visual vector mapped to the language space be *closer* to the correct answer than to any of the distractors. This can be achieved by learning a mapping $\mathcal{M}$ that projects a video $x$ to the language embedding space by minimizing the triplet metric learning loss used in [35], i.e.,:

$$\mathbf{W}^* = \operatorname*{argmin}_{\mathbf{W}} \sum_{i=1}^{n} \sum_{t \neq t^i} \left[ \|\mathcal{M}(x^i; \mathbf{W}) - \phi_l(y_t^i)\|_2^2 - \|\mathcal{M}(x^i; \mathbf{W}) - \phi_l(y_{t^i}^i)\|_2^2 + \alpha \right]_+. \quad (1)$$

$\mathcal{M}(x; \mathbf{W})$ is a mapping with parameters $\mathbf{W}$. $\mathcal{M}(x; \mathbf{W})$ can be a deep ConvNet trained on raw input video $x$. In a simpler case, it takes a predefined $\phi_v(x)$ as input and applies a simple linear projection. $[.]_+$ is the hinge function. Finally, $\alpha$ is a hyper-parameter to control the margin between the distance to the true sentence $y_{t^i}^i$ and the distances to the wrong sentences $y_t^i$ (with $t \neq t^i$). At test time, this approach chooses the CC that is closest to the predicted vector $\mathcal{M}(x; \mathbf{W})$ in the language embedding space: $t^* = \operatorname{argmin}_{t \in [1..k]} \|\mathcal{M}(x; \mathbf{W}^*) - \phi_l(y_t)\|_2^2$.

## 5. Experiments

### 5.1. Experimental setup

**Language embedding**: For all the experiments in this paper, we use word2vec [28] as the language embedding $\phi_l(y)$. word2vec is a shallow network trained on a large corpus to reconstruct the linguistic context of the words. It has been shown that this word embedding can map words having similar contexts into language vectors that are close. We use the word2vec model provided by [28], which is pre-trained on the Google News dataset. This gives a 300-dimensional vector for each word. To represent $\phi_l(y)$, we extract word2vec vectors for all words in $y$, average these vectors, then L2-normalize the average vector to build a language representation for the sentence $y$. Appendix B contains experiments based on other language embedding models, including skip-thought vectors [21], hidden states of an encoder-decoder LSTM trained for machine translation [27], as well as FastText features [9].

**Visual embedding**: We use different visual representations for $\phi_v(x)$ in different experiments. These representations are computed from different ConvNet architectures pre-trained on different datasets. We feed frames (or clips in the case of C3D) into these pre-trained ConvNets to extract activations of a particular layer and use them as representations. We use the VGG network [37], and C3D [43]. The pre-trained models are provided by the authors of [37, 43]. For simplicity, from now on we denote these representations as VGG, and C3D, respectively. We specify a visual representation by a pair of an architecture name and a layer name, e.g., VGG-fc6. It is worth noting that videos in ViCom have varying length (from a few dozens to a few hundreds frames). While VGG operates on frames, C3D takes as input short clips of 16 frames. In order to build a fixed-size visual representation for a video, for the case of VGG we average the frame features over the entire video, and then L2-normalize the average vector. For the case of C3D we average the clip features extracted from all 16-frame clips in the video and L2-normalize the resulting vector.

Because the task of Video Multiple Choice Caption requires mapping from video to text, we also consider a visual embedding obtained from a video captioning model that has been explicitly trained on this dual source of data. For this purpose we use the S2VT captioning model described in [44] which was trained on the MSVD dataset [10]. S2VT is a 2-hidden-layer LSTM that takes as input a video (represented in the form of the sequence of VGG-fc7 frame features) and predicts a text sentence. The model is an encoder-decoder, where the hidden state from the first LSTM can be interpreted as the encoded video and the second LSTM is optimized to decode this representation into text. Thus, we use S2VT as visual embedding by taking the hidden state of the first LSTM after having fed the entire video as input. We denote these visual features as S2VT.

**Regression models**: We experimented with linear regression applied to C3D-fc6, VGG-fc6 and S2VT. We call these approaches **LR-C3D**, **LR-VGG**, and **LR-S2VT**, respectively. However, we found that directly regressing the average word2vec representation $\phi_l(y)$ of the correct caption led to very poor results with all visual features. We attributed this phenomenon to the high-dimensionality (300) of the target vector $\phi_l(y)$ and the overly-harsh objective of regression which aggressively attempts to map each training visual embedding $\phi_v(x^i)$ as close as possible to its ground truth language embedding $\phi_l(y_t^i)$. We found empirically beneficial to reduce the dimensionality of the $\phi_l(y_t^i)$ using Principal Component Analysis (PCA) before training the regressors. Results are reported using 30 PCA dimensions, which was found to be the optimal dimensionality on the validation set.

We have also attempted to train a modified version of C3D as a deep regressor. The layers of this model are identical to those of C3D up to fc6. We then added a fully-connected layer with linear activation to linearly project into the PCA-ed language embedding space $\phi_l(y)$. However, even when using heavy regularization via weight decay, we found the results achieved with this deep regressor to be inferior to those obtained with simple linear regression.

| Approach | Method | Features | Training | Acc (%) |
|---|---|---|---|---|
| Ignore video input | Random | - | - | 20.0 |
| | Shortest CC | - | - | 13.0 |
| | Median CC | - | - | 23.0 |
| | Longest CC | - | - | 20.3 |
| Regression | LR-VGG | VGG-fc6 | - | 24.0 |
| | LR-C3D | C3D-fc6 | - | 26.1 |
| | LR-S2VT | S2VT | - | 23.7 |
| Metric learning | SML-VGG | VGG-fc6 | - | 51.0 |
| | SML-C3D | C3D-fc6 | - | 51.2 |
| | SML-S2VT | S2VT | - | 42.8 |
| | SML-Comb | Comb-fc6 | - | 51.6 |
| | DML-C3D-0 | - | scratch | 45.6 |
| | DML-C3D-F | - | finetuned | **53.9** |

Table 2: **Accuracy of several baseline models on the complete VideoMCC test set** (81, 476 **clips**). Methods based on metric learning give the best performance. Random chance is at 20%.

**Metric learning models**: We experiment with two different sets of architectures: shallow and deep networks. In the shallow network setting, we assume that we have a reasonably good visual representation, and we just learn a single fully-connected layer without nonlinear unit. We optimize this model by the triplet loss (as described in section 4.2). We test the shallow metric learning with three different representations: VGG-fc6, C3D-fc6, and S2VT. We name these approaches **SML-VGG**, **SML-C3D**, and **SML-S2VT**, respectively. We also test this shallow net applied to a combined representation of VGG-fc6 and C3D-fc6 (a simple concatenation). We name this approach **SML-Comb**. For the deep network setting, we use again an architecture similar to C3D. We use all layers identical to those of C3D up to fc6. We then add a linear fully-connected layer with 300 output units to match the dimensionality of word2vec. We can either train this network from scratch or finetune it from C3D. We name these approaches **DML-C3D-0** and **DML-C3D-F**, respectively.

**Training settings**: Both shallow and deep networks are trained using SGD with a momentum of 0.9. For shallow networks, we use a mini-batch size of 128. The initial learning rate is 0.01 and it is reduced by a factor of 0.1 every 10K iterations. Training is stopped at 60K iterations. For deep networks, we use a mini-batch size of 30. The initial learning rate is $3 \times 10^{-5}$ for DML-C3D-0 and DML-C3D-F. It is reduced by 0.1 for every 100K iterations and the training is stopped at 300K iterations. Since training deep networks is time-consuming, we choose $\alpha$ by cross validation on shallow networks. Our experiments show that using $\alpha = 0.1$ gives the best results among the tested margins of 0.01, 0.1, and 1 for all visual features features. Thus, we use $\alpha = 0.1$ in all deep metric learning networks.

## 5.2. Experimental results

**Evaluation on VideoMCC**. Table 2 presents the accuracy of our different approaches on VideoMCC. We also include simple reference baselines that ignore the video input and choose CCs either randomly or according to their lengths (e.g., always the shortest, the longest, or the one with median length among the set of five CCs). Among the regression methods, LR-C3D performs the best, but it is just 6.1% better than random chance. All metric learning models perform much better than regression. This suggests that directly regressing the visual embedding into the language embedding is an overly aggressive learning objective. Shallow metric learning methods perform quite well for all visual representations, and SML-C3D gives the best performance (51.2%) among the shallow networks with a single representation. Combining visual representations boosts the accuracy to 51.6%. The fine-tuned deep network DML-C3D-F gives the highest accuracy (53.9%).

We have also tested several additional variants of the proposed methods. For example, we trained shallow metric learning models with respect to the cosine distance and the dot-product but these variants produced either no change or inferior performance: e.g., 46.5% and 50.1%, respectively, for the case of SML-VGG. We experimented also with a version of SML-VGG where instead of averaging the features over all frames, we used the VGG−fc6 features extracted from the central frame of the video. In the case of single frame, the performance of SML-VGG drops to 42.3% from the accuracy of 51.0% achieved by averaging the frame features over the entire video. This suggests that VideoMCC tests can be more accurately solved by an analysis of the entire clip as opposed a single frame.

**Human study**. We also conducted a human study in order to gauge how the algorithms compare to human performance on this benchmark. For this study, a subset of 8,733 clips were randomly drawn from the test split. Each clip and its associated multiple-choice test were shown (without audio) to human annotators on the Amazon Mechanical Turk (AMT) platform [1]. We asked the annotators to select the sentence that best describes the video clip out of the $k = 5$ choices. For each clip, we collected a total of 5 selections from different AMT annotators. Because TV news often contain overlaid tickers with informative text summarizing the news, we decided to obscure the tickers via blurring, so that AMT workers were forced to choose the CC purely based on the content of the video (as the machine) rather than by reading the tickers. To obscure the tickers, we trained an SVM text detector using HOG features [12] and then heavily blurred the portion of the image containing the detected ticker.

Table 3 reports the comprehension accuracy for humans and the different algorithms on this subset of 8,733 clips (note that since this study is performed on a subset of the test set, the accuracy results differ slightly from those reported in Table 2). Humans achieve an accuracy of 63.4%. This indicates that VideoMCC is a hard task and that our best approach, DML-C3D-F, is still 7.5% below human-level performance. The table shows also accuracy by topic. Humans outperform machines across all topics, but DML-C3D-F nearly approaches human-level performance on a few categories, such as Technology and Crime. DML-C3D-F performs reasonably well across all topics. LR-C3D is clearly inferior on all categories and does worse than random chance on the topic of Emotion.

**Validation on strongly-supervised test set**. The far-from-flawless performance achieved by humans on our benchmark outlines the difficulty of the task. At the same time one may argue that a portion of this error may actually be due to noisy labels in the dataset, e.g., caused by by ambiguous captions or by CCs being not perfectly aligned with the video. In order to assess the impact of such issues on the accuracy measured by our benchmark, we used the AMT annotations described above to produce a smaller, strongly-supervised version of the test set. Specifically, we formed a "filtered" test set by considering only the video clips where the ground truth CC was chosen 3 or more times (out of 5) by the AMT workers. This gave us a total of 5,923 strongly-labeled video clips out of the original set of 8,733 clips analyzed by the AMT workers. Table 4 summarizes the results achieved by our algorithms on this strongly-supervised test set (Filtered set) versus the superset of 8,733 clips (Unfiltered set), which contains also the clips that caused 3 or more errors out of 5 annotations. If we compare the 2 columns in this table, we see that the accuracies on the filtered set are higher for all methods. This makes sense, since the refined test set includes only videos where the majority of the AMT workers chose the ground truth CC. These videos are less ambiguous for humans compared to those that were filtered out. Thus, it is expected that also machines perform better on this subset compared to the unfiltered set.

However, the most important observation to draw from this study is that, while the *absolute* performance of the algorithms on the filtered set is different from that measured on the complete set, their *relative* performance (and rank order) is unchanged. This suggests that our original test set is a good benchmark to assess relative performance of algorithms, despite having been constructed without human validation. Note that, while validating the entire test set may provide more accurate absolute numbers, it would constrain the size of the benchmark to be very small due to the huge human cost involved. Conversely, our semi-automatic procedure requires minimal human intervention and as such it can be leveraged to create datasets of arbitrary scale with fixed human cost: the only human-annotations needed in our approach are the labels for the relevant-clip detector, and these annotations are independent of the benchmark size.

| Topic | Politics | Climate | Election | Time | Misc | Tech | Legal | Economics | Crime | Emotion | All |
|---|---|---|---|---|---|---|---|---|---|---|---|
| LR-C3D | 30.4 | 32.5 | 31.4 | 21.2 | 25.6 | 26.1 | 29.3 | 31.8 | 32.5 | <u>15.3</u> | 27.5 |
| DML-C3D-F | 60.5 | 64 | 60.8 | 47.9 | 45.7 | 60 | 49.2 | 58.6 | 63.3 | 47.3 | 55.9 |
| Humans | **64.9** | **70.9** | **65.4** | **58.3** | **58.6** | **66.1** | **56.1** | **61.3** | **70.5** | **57.9** | **63.4** |

Table 3: **VideoMCC performance by humans and machines on a subset of 8,733 test clips validated with AMT**. Performance is shown by topic as well as for all categories (All). Humans achieve the highest accuracy across all topics. Accuracy below random chance (20%) is shown in <u>underlined</u> text. DML-C3D-F attains results close to human-level performance on Climate, Technology and Crime.

| Method | Accuracy (%) | |
|---|---|---|
| | Unfiltered subset (8,733 clips) | Filtered subset (5,923 clips) |
| LR-VGG | 25.8 | 29.4 |
| LR-C3D | 27.5 | 31.5 |
| LR-S2VT | 25.5 | 28.7 |
| SML-VGG | 52.6 | 60.6 |
| SML-C3D | 53.1 | 61.5 |
| SML-S2VT | 46.3 | 53.3 |
| SML-Comb | 52.8 | 61.6 |
| DML-C3D-0 | 47.9 | 54.6 |
| DML-C3D-F | 55.9 | 64.9 |

Table 4: **Accuracy on a strongly-supervised subset of the VideoMCC test set**. From a total of 8,733 clips assessed by AMT workers (Unfiltered subset), we built a strongly-supervised subset of 5,923 clips (Filtered set). While the accuracy for all methods is higher on the filtered set compared to the unfiltered set, the *relative* performance (and rank order) of the algorithms remains unchanged on the two sets. This suggests that, despite not using any human labeling, our original test set is a good benchmark to assess relative performance of algorithms.

| **Error reason** | **Percentage (%)** |
|---|---|
| Video and ground truth CC are weakly related | 43.1 |
| Distractor CC is as good or better | 20.8 |
| Video is advertisement | 5.3 |
| In-studio clip with news anchors | 2.8 |
| Clip contains static scene | 6.4 |
| Ground truth CC seems obvious choice | 14.8 |

Table 5: **Analysis of human errors on VideoMCC**. We visually inspected 283 clips where the ground truth CC was selected 2 or fewer times (out of 5) by AMT workers. We manually diagnosed the source of the error into the six categories listed in this table.

Finally, in order to gain further insights about the reasons of the human errors, we visually inspected a total of 283 test videos randomly chosen from those where the ground truth CC was selected only 2 or fewer times (out of 5) by the AMT workers. We manually labeled each video with six possible failure reasons, as outlined in Table 5. In 43.1% of the error cases, the human errors seem caused by a disconnect between the semantic content of the videos and the CCs. We view this as errors that are intrinsic to the task and due to semantic ambiguities in either the video or the ground truth CC, rather than being caused by failures in our system. Conversely, rows 2 through 5 in Table 5 list error cases that could in principle be eliminated by a better relevant clip detector (rows 3 through 5) or by a procedure that chooses distractors dissimilar to the ground truth (row 2). We plan to investigate such improvements in future work. Finally, in 14.8% of the error cases, we found the ground truth CC to be the obvious choice and thus we could not provide an explanation for the error in the AMT selections, except for the inherent noise present in AMT annotations.

## 6. Conclusions

In this paper we introduced a new form of video understanding assessement, we presented a general procedure to construct semi-automatically benchmarks for this task, we created a dataset that we plan to release to the community and we evaluated a series of approaches on it. VideoMCC fulfills the following desirable properties: 1) it defines a well-posed task with a good quantitative evaluation metric; 2) it assesses the ability to semantically comprehend video; 3) it is large-scale, thus enabling effective training of deep models. We hope that this new task and our benchmark will become useful stepping stones to fundamentally transform video analysis into higher-level video understanding. We have seen this happening in the image domain where a new large-scale benchmark [34] married with a powerful machine learning model [23] gave rise to a new generation of computer vision algorithms. We also expect that our benchmark will spur active research at the intersection between video understanding and natural language processing.

Future work will be devoted to improving the clip detector by leveraging both the CCs and the video in order to address some of the errors identified in our analysis. As our dataset construction is semi-automatic we believe that it will be possible to scale up VideoMCC quickly to a much larger benchmark with little human, computational and financial cost. We expect to increase the dataset by an order of magnitude within the next year. We will also experiment with more sophisticated approaches to generate distractor CCs in order to make the task even more challenging. In order to stimulate steady progress in this area, we plan to organize a series of grand challenges built around our benchmark. The community will help us raise the level of the state-of-the-art on our benchmark in the future. We will release the VideoMCC dataset, all implementations and models upon publication of this article.

**Acknowledgment**: We would like to thank Dimitrios

Latsis, Trevor von Stein and Roger Macdonald from Internet Archive for providing raw videos and closed captions for the creation of the dataset. We are grateful to Mark Williams for helpful discussions. We thank Subhashini Venugopalan for providing S2VT code, pre-trained models, and clarifications about the S2VT parameter choices. This work was funded in part by Facebook AI Research and NSF award CNS-1205521. We gratefully acknowledge NVIDIA for the donation of GPUs used for portions of this work.

# Appendices

## A. VideoMCC topics and examples

In the paper we presented an experiment where we use LDA [7] to model topics of VideoMCC sentences using 10 topics. The top words for each topic are presented in Table 6.

Figure 3 shows some video examples from the VideoMCC dataset with their corresponding ground truth sentences. It can be noticed that the examples cover a wide array of subjects ranging from environmental events, to science, politics, and accidents. This renders VideoMCC a good benchmark in terms of content diversity.

In our human study we found that weak semantic correlation between the video and the ground truth CCs was one of the sources of error (see Table 5). In Figure 4 we present some illustrative examples of this type of error.

## B. Additional experiments

### B.1 Language embedding models

The experiments presented in §5.2 are obtained by using average `word2vec` vectors as representation for the sentences in the multiple-choice tests. Here we consider other language embedding models and measure their efficacy on our benchmark.

We experiment with the following language embedding models: `skip-thought` vectors [21], hidden states of an encoder-decoder LSTM trained for machine translation (Sequence-to-Sequence encoder) [27], and FastText features [9]. Skip-thought is a sentence-to-vector model mapping a sentence to a descriptor of $4,800$ dimensions trained on the BookCorpus dataset [49]. Sequence-to-Sequence (S2S) is a multi-layer LSTM model trained to translate from one language to another. Our implementation is based on the pre-trained model provided in [17] which is optimized to translate from English to German. We use the 4-layer LSTM model with 1000 hidden units. We encode each sentence with this model and use the values of the 1000 hidden units as sentence representation. FastText is a language embedding based on the skip-gram model trained on the large-scale YFCC100M corpus [40]. This embedding has 300 dimensions.

We retrained our shallow metric learning networks (**SML-VGG**, **SML-C3D**, and **SML-Comb**) on each of these different sentence representations (instead of the average `word2vec`). Table 7 shows that, among these alternative language representations, `skip-thought` produces the strongest performance. It even outperforms `word2vec`.

### B.2 Relevant-clip detection

Figure 5 presents the ROC curve of our relevant-clip detector evaluated on the validation set. The area under the ROC curve is $0.94$. We select the threshold cut-off at the point where the false positive rate is $0.1$. At this cut-off point the true positive rate is $0.83$. This gives a good trade-off between recall and precision.

## C. Additional validation on strongly supervised test set

In addition to model performance on the strongly supervised subset of the test set reported in §5.2, here we provide accuracy obtained with more stringent conditions of human agreement (4+ and 5 correct answers from 5 AMT annotators) on filtered sets. The results are shown in the Table 8. It can be seen that accuracy for all methods becomes higher as we apply stricter conditions of human agreement. Note that the *relative* performance of the algorithms remains largely unchanged on all sets, which suggests that even the unfiltered set is a good benchmark to judge relative performance.

In order to assess human performance on a strongly supervised test set we collected additional AMT annotations for a subset of 1000 clips out of the total $5,923$ videos in a filtered subset. The resulting accuracy is $84.9\%$ as opposed to $63.4\%$ achieved by the AMT workers on the unfiltered test set.

## D. Implementation details

We discuss here the details of our training procedure and the hyper-parameter values used for both shallow (SML) and deep networks (DML).

Both shallow and deep networks are trained using SGD with a momentum of $0.9$. For shallow networks, we use a mini-batch size of $128$. The initial learning rate is $0.01$ and it is reduced by a factor of $0.1$ every 10K iterations. Training is stopped at 60K iterations.

For deep networks, we use a mini-batch size of $30$. The initial learning rate is $3 \times 10^{-5}$ both for DML-C3D-0 and DML-C3D-F. It is reduced by $0.1$ for every 100K iterations and the training is stopped at 300K iterations. Since training deep networks is time-consuming, we choose $\alpha$ by cross validation on shallow networks. We found in our experiments that $\alpha = 0.1$ gives the best results among the tested margins of $0.01$, $0.1$, and $1$ for all visual features features. Thus, we use $\alpha = 0.1$ in all deep metric learning networks.

| Topic | Top words |
|---|---|
| Politics | `government, problem, country, american, war, military, protest, attack, unit, security, question, force, leader, nation, call` |
| Climate | `area, city, storm, fire, hour, water, north, across, air, mile, center, snow, force, south, weather, power, rain, thousand, hit, through` |
| Election | `house, republican, big, obama, romney, white, senate, democrat, vote, last, campaign, election, party, game, mitt, governor, win, night, race, poll` |
| Time | `next, tonight, story, world, hour, weekend, around, few, numberth, week, ahead, show, begin, america, chuck, start, stay, york, daily` |
| Technology | `san, kill, old, man, francisco, west, future, police, cover, bloomberg, shot, business, technology, pier, welcome, men, hospital` |
| Legal | `case, court, call, charge, investigate, police, law, response, release, decision, office, official, former, action, death, against, depart, defense` |
| Economics | `dollar, million, care, job, health, cut, tax, plan, than, paid, money, program, billion, government, american, company, announcement, raise, develop` |
| Crime | `close, car, police, street, off, school, fire, video, inside, scene, show, build, last, crash, worker, open, wall, park, home, office` |
| Emotion | `thank, much, little, let, join, learn, way, washington, stephanie, love, read, early` |
| Miscellaneous | `thing, them, lot, put, because, really, work, got, happen, did, keep, these, very, something, way, try, need, well, any` |

Table 6: **VideoMCC topics**. Topic modeling of VideoMCC sentences using LDA [7] with 10 topics. The topic titles were manually chosen by us based on inspection of the top words.

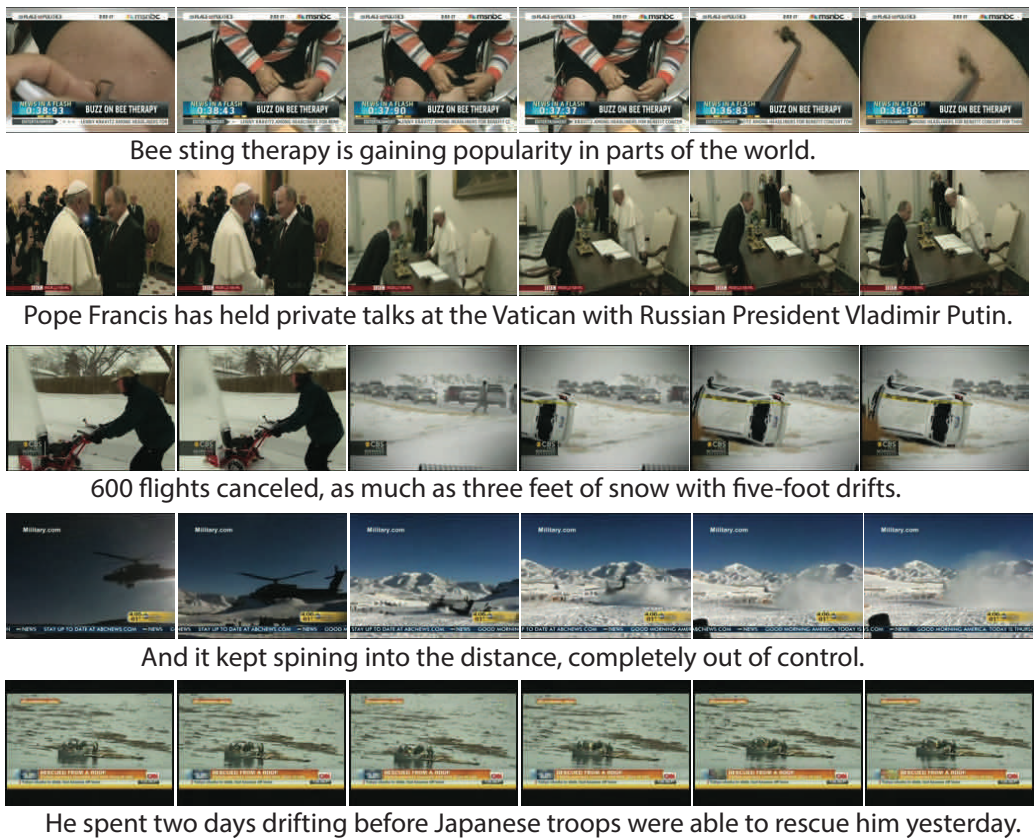

Bee sting therapy is gaining popularity in parts of the world.

Pope Francis has held private talks at the Vatican with Russian President Vladimir Putin.

600 flights canceled, as much as three feet of snow with five-foot drifts.

And it kept spining into the distance, completely out of control.

He spent two days drifting before Japanese troops were able to rescue him yesterday.

Figure 3: **VideoMCC examples**. Some video clips from the VideoMCC dataset with their ground truth closed caption sentences. Since our benchmark is created from news videos, it naturally covers a wide array of subjects ranging from environmental events, to science, politics, and accidents.

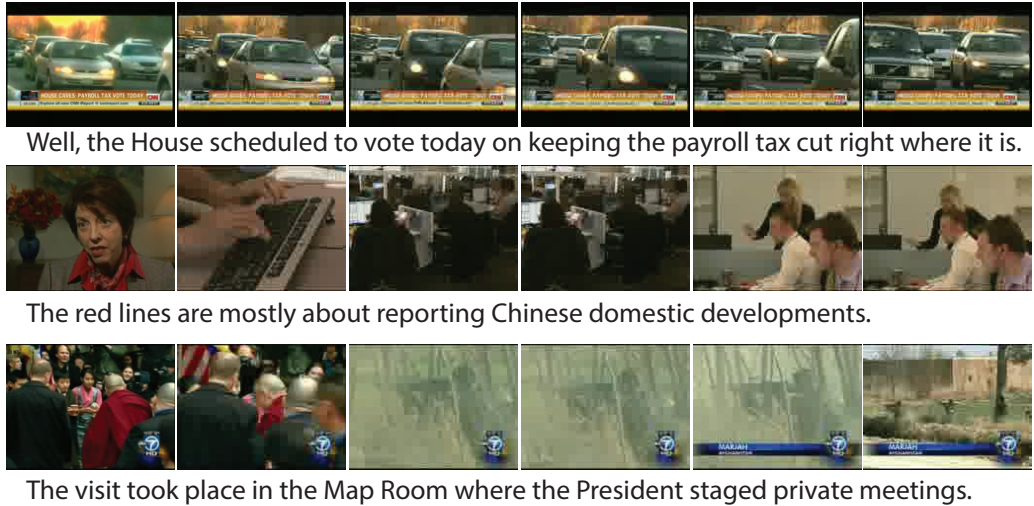

Well, the House scheduled to vote today on keeping the payroll tax cut right where it is.

The red lines are mostly about reporting Chinese domestic developments.

The visit took place in the Map Room where the President staged private meetings.

Figure 4: **Examples of VideoMCC with weak connection between CCs and visual cues**. Some video clips from VideoMCC dataset with their ground truth closed caption sentences weakly related to visual content of the clip.

| $\phi_l$ option | $\phi_l$-dim | SML-VGG | SML-C3D | SML-S2VT | SML-Comb |
|---|---|---|---|---|---|
| word2vec | 300 | 51.0 | 51.2 | 42.8 | 51.6 |
| skip-thought | 4,800 | 56.9 | 57.7 | 49.0 | 59.96 |
| S2S-encoder | 1000 | 41.1 | 42.3 | 36.3 | 42.8 |
| FastText | 300 | 46.7 | 47.5 | 40.1 | 48.8 |

Table 7: **Experiments with different language models**. VideoMCC accuracy of shallow metric learning networks using different sentence representations $\phi_l$.

| Method | Accuracy (%) | | | |
|---|---|---|---|---|
| | Unfiltered (8,733 clips) | Filt. (3+) (5,923) | Filt. (4+) (4,364) | Filt. 5 (2,335) |
| LR-VGG | 25.8 | 29.4 | 30.7 | 33.94 |
| LR-C3D | 27.5 | 31.5 | 32.9 | 36.5 |
| LR-S2VT | 25.5 | 28.7 | 29.9 | 32.6 |
| SML-VGG | 52.6 | 60.6 | 64.8 | 69.4 |
| SML-C3D | 53.1 | 61.5 | 65.3 | 68.7 |
| SML-S2VT | 46.3 | 53.3 | 56.6 | 61.5 |
| SML-Comb | 52.8 | 61.1 | 65 | 68.9 |
| DML-C3D-0 | 47.9 | 54.6 | 57.9 | 61.1 |
| DML-C3D-F | 55.9 | 64.9 | 68.5 | 71.9 |

Table 8: **Accuracy on strongly supervised subsets of the VideoMCC test set.** From a total of 8,733 clips annotated by AMT workers (Unfiltered set), we built strongly supervised subsets of 5,923, 4,364 and 2,335 clips corresponding to multiple-choice tests that received 3 or more, 4 or more, and 5 correct answers from annotators, respectively.

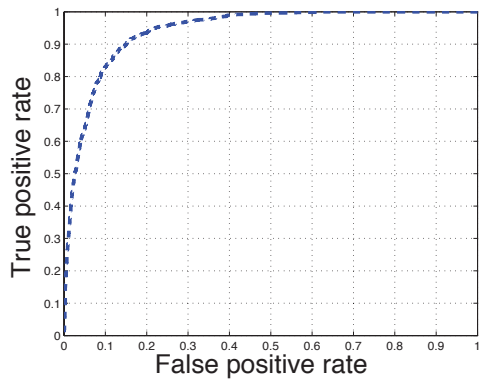

Figure 5: **Relevant clip detection**. The ROC curve of our relevant clip detector evaluated on the validation set. At the false positive rate of 0.1, the true positive rate is 0.83. The area under the ROC curve is 0.94.